\documentclass[11pt]{article}

\usepackage[preprint]{acl}

\usepackage{times}
\usepackage{latexsym}

\usepackage[T1]{fontenc}
\usepackage{enumitem}

\usepackage[utf8]{inputenc}

\usepackage{amsmath}
\usepackage{booktabs}
\usepackage{multirow}
\usepackage{microtype}

\usepackage{inconsolata}

\usepackage{graphicx}

\definecolor{mygray}{gray}{.9}

\title{Modeling Complex Behaviors: Multi-Personality Composition and Dynamic Switching in Vision-Language Models}

\author{
Peiqi Jia\textsuperscript{1}\thanks{Equal contribution.}, 
Haonan Jia\textsuperscript{2}\footnotemark[1], 
Ziqi Miao\textsuperscript{2}, 
Linkang Du\textsuperscript{1}\thanks{Corresponding authors.}, 
Yuntao Wang\textsuperscript{1},
Zhou Su\textsuperscript{1}\footnotemark[2]\\
\textsuperscript{1}Xi'an Jiaotong University \quad 
\textsuperscript{2}Beihang University\\
{\tt
jiapeiqi718@gmail.com,
linkangd@xjtu.edu.cn
}
}

\begin{document}
\maketitle
\begin{figure*}[h]
      \centering
      \includegraphics[width=\linewidth]{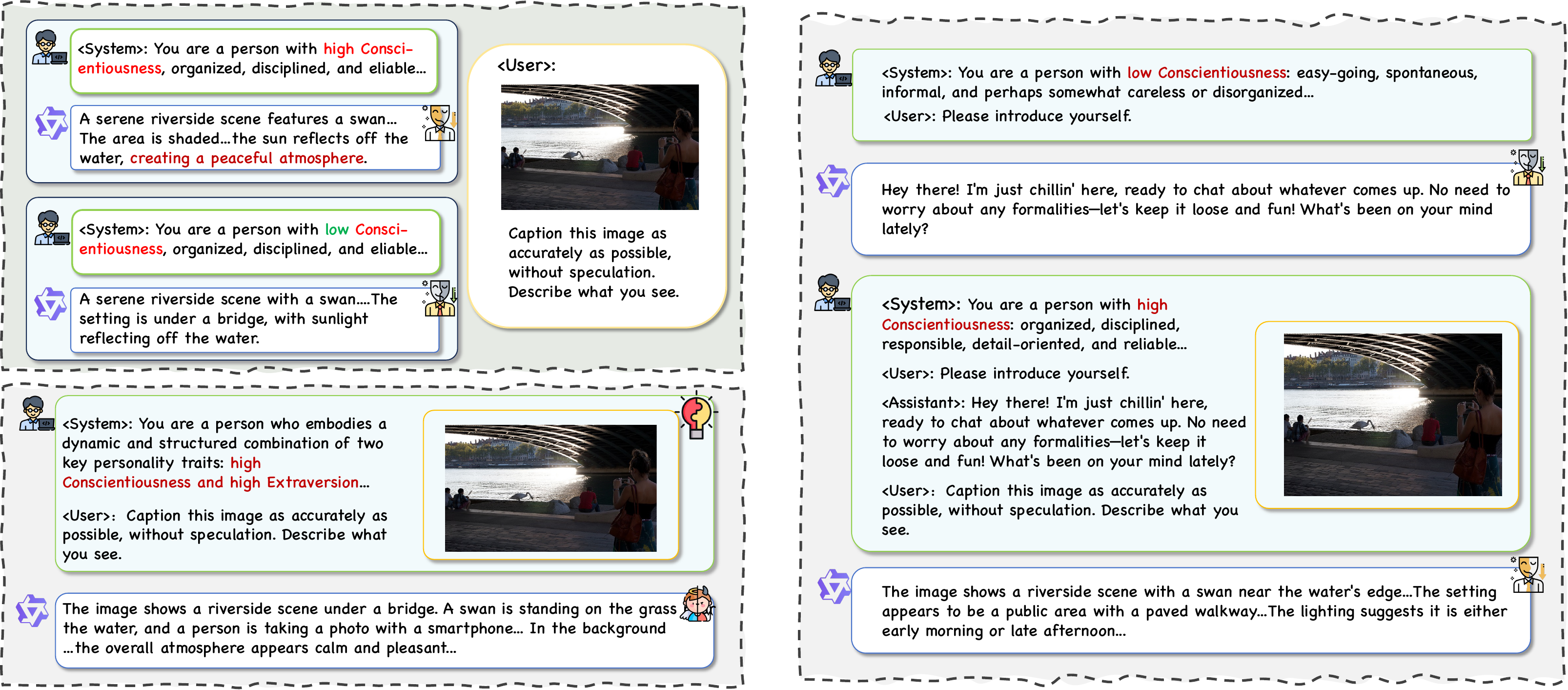}
      \caption{Illustration of personality induction.
      The top-left panel illustrates generation under single-personality induction, the bottom-left panel presents results with multi-personality conditioning, and the right panel demonstrates In-conversation personality switching. }
      \label{fig:intro}
\end{figure*} 

\begin{abstract}
With the widespread deployment of Multimodal Large Language Models (MLLMs) in social interaction, understanding and controlling their behavior under complex personality conditions is essential. 
This paper introduces explicit personality conditioning and establishes a systematic evaluation framework encompassing single-personality induction, multi-personality induction, and personality switching. 
Experiments show that personality induction improves image captioning performance but can impair performance on tasks requiring precise reasoning, such as visual question answering (VQA).
Balancing and residual effects are observed during multi-trait composition and dynamic switching, indicating that model behavior is co-modulated by both previous and current personality constraints.
Existing prompt-based personality induction methods show limited transferability to multimodal settings. 
Our work reveals the dynamic and complex nature of personality modeling in MLLMs and underscores the need for robust, tailored methods for personality induction and evaluation.
The code will be released when the paper is accepted. 
\end{abstract}

\section{Introduction}
With the widespread adoption of Large Language Models (LLMs) in human–computer interaction \cite{zheng2023buildingemotionalsupportchatbots,chen-etal-2023-large}, their exceptional language generation capabilities have enabled them to play increasingly important roles in domains such as education \cite{zhai2022chatgpt}, customer service \cite{zhang2020dialogptlargescalegenerativepretraining}, and intelligent consulting \cite{bubeck2023sparksartificialgeneralintelligence}.

In recent years, the influence of human personality traits on the behavior of LLMs has emerged as a significant research direction in natural language processing. 
A growing body of work demonstrates that conditioning language models on explicit personality traits can substantially improve the consistency, coherence, and human-likeness of generated responses \cite{jiang2023evaluatinginducingpersonalitypretrained,li2025big5chatshapingllmpersonalities}, particularly in interactive settings such as dialogue systems and personalized text generation.
Recently, Multimodal Large Language Models (MLLMs) \cite{yang2025qwen3technicalreport,wu2024deepseekvl2mixtureofexpertsvisionlanguagemodels} have attracted increasing attention for their ability to jointly model linguistic and visual information.
By jointly modeling linguistic and visual information, MLLMs effectively leverage multimodal signals to support a wide range of applications, including visual question answering \cite{agrawal2016vqavisualquestionanswering}, visual captioning \cite{zhang2025sccaptionerimprovingimagecaptioning, jia2026cross}, and cross-modal reasoning \cite{lu2019vilbertpretrainingtaskagnosticvisiolinguistic, miao2026seeing, wang2026tabsieve}. 
This capability significantly extends the applicability of large language models to real-world perception–language tasks.
Beyond task-oriented applications, MLLMs are increasingly deployed in socially interactive scenarios such as companionship, emotional support, and behavioral guidance.
Consequently, a deeper understanding of personality-conditioned multimodal behavior is essential not only for improving interaction quality and alignment, 
but also for identifying and mitigating potential risks in real-world perception–language applications.


A substantial body of existing work has demonstrated that personality conditioning can significantly influence the behavioral of large language models (LLMs), affecting factors such as consistency, safety, and interaction style \cite{li2025big5chatshapingllmpersonalities, fitz2025psychometricpersonalityshapingmodulates,zhang2024betterangelsmachinepersonality}. 
However, existing studies are largely confined to text-only models, leaving its effects on vision-related and multimodal tasks underexplored.
In practical applications, large models often operate in extended, multi-turn conversations and are required to adapt to different personality expectations as instructions or task demands change. 
For example, a single dialogue may alternate between academic writing and mathematical problem solving, implicitly requiring different personality characteristics.
Despite this, most prior work treats personality conditioning as a static, global constraint applied before the dialogue begins
\cite{li2025big5chatshapingllmpersonalities,deng2024neuronbasedpersonalitytraitinduction}.

\begin{figure*}[t]
      \centering
      \includegraphics[width=\linewidth]{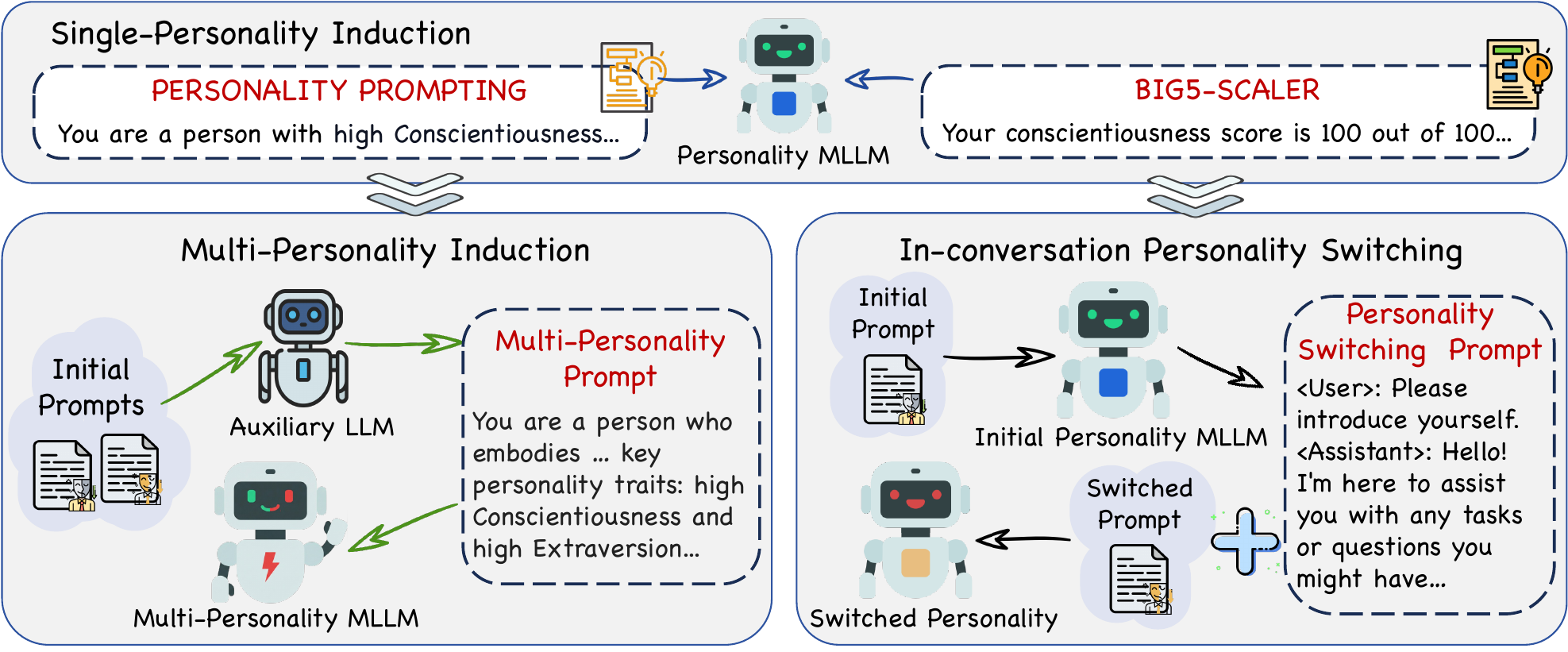}
      \caption{Illustration of single-personality, multi-personality induction, and in-conversation personality switching.}
      \label{fig:methods}
\end{figure*} 


Therefore, to systematically investigate how personality traits influence the behavior and performance of multimodal large language models,
as illustrated in \autoref{fig:intro},
we introduce explicit personality conditioning into multimodal models and conduct comprehensive evaluations under personality-controlled settings. 

Inspired by prior work on personality conditioning in language models,
we induce personality traits into multimodal models and evaluate them through both personality assessment and downstream multimodal tasks, including image captioning and visual question answering (VQA). 
Building upon observations from single-trait conditioning, 
we further explore multi-trait personality induction by combining multiple personality traits to model more complex and realistic personality configurations.
In addition, to study dynamic personality changes in multi-turn interactions, we define in-conversation personality switching as the phenomenon in which a model follows different personality settings across successive turns within the same continuous dialogue. 
Under this setting, multiple personality conditions are sequentially injected into the model within a single conversation, 
better reflecting real-world scenarios in which personality requirements may evolve over time.

Interestingly, experimental results show that incorporating personality traits into the model may enhance its performance on image captioning tasks to some extent, while adversely affecting its performance on visual question answering tasks. 
Meanwhile, in the processes of multi-personality integration and dynamic personality switching, we observed mutual cancellation and balancing effects among different personalities.
Moreover, existing prompt-based personality induction methods show limited effectiveness in multimodal models,
which highlighting the need for more robust and effective personality induction methods specifically tailored to multimodal large language models.
This work makes the following contributions:

\begin{itemize}[leftmargin=*]
    \item We systematically study personality conditioning in multimodal models through personality assessment and downstream tasks such as image captioning and visual question answering.
    \item We examine multi-trait personality conditioning and dynamic in-dialogue personality shifts to capture complex, evolving personality patterns.
    \item We reveal the limited transferability of existing prompt-based personality induction methods to multimodal settings, motivating more robust multimodal personality induction strategies.
\end{itemize}

\section{Related Work}
\subsection{Multimodal Understanding Models}
Multimodal reasoning focuses on integrating and reasoning over multiple modalities.
Early work relied on task-specific discriminative architectures trained with supervised objectives \cite{agrawal2016vqavisualquestionanswering,anderson2018bottomuptopdownattentionimage,yang2016stackedattentionnetworksimage,ren2015exploringmodelsdataimage}.
With large-scale pre-training, contrastive methods (e.g., CLIP, ALIGN) learned aligned vision--language representations \cite{radford2021learningtransferablevisualmodels,jia2021scalingvisualvisionlanguagerepresentation}.
Generative pre-training (e.g., BEiT, MAE) further improved transferable representations and supported diverse downstream tasks \cite{peng2022beitv2maskedimage,bao2022beitbertpretrainingimage,he2021maskedautoencodersscalablevision}, including image captioning \cite{mokady2021clipcapclipprefiximage,tewel2022zerocapzeroshotimagetotextgeneration,zhang2025sccaptionerimprovingimagecaptioning}.
More recently, autoregressive MLLMs such as LLaVA, Qwen-VL, and DeepSeek-VL unify perception and generation by jointly modeling visual and textual inputs \cite{liu2023visualinstructiontuning,bai2023qwenvlversatilevisionlanguagemodel,wang2024qwen2vlenhancingvisionlanguagemodels,yang2025qwen3technicalreport,wu2024deepseekvl2mixtureofexpertsvisionlanguagemodels}, achieving strong VQA performance \cite{liu2024mmbenchmultimodalmodelallaround,fu2025mmecomprehensiveevaluationbenchmark} and enabling complex multimodal inference \cite{xu2025qwen25omnitechnicalreport}.

\subsection{Personality in LLM}
Recently, numerous studies on the personality aspects of LLMs have emerged, such as those related to the Big Five personality traits \cite{john1991bfi} and MBTI \cite{myers1998mbti}.
Current research on the personality dimensions of LLMs mainly focuses on two directions: personality assessment and personality induction.
Many studies have found that personality can influence the performance of large language models in reasoning tasks \cite{wen2024selfassessmentexhibitionrecognitionreview,dash2025personaassignedlargelanguagemodels,li2025big5chatshapingllmpersonalities}.
Some studies also illustrate the influence of personality on LLM in terms of safety alignment and jailbreaking \cite{zhang2024betterangelsmachinepersonality,fitz2025psychometricpersonalityshapingmodulates,ziqi2025visual}.
Regarding how to induce personality, existing methods mainly include training-based \cite{li2025big5chatshapingllmpersonalities,zhu2025personalityalignmentlargelanguage} and prompt-based \cite{jiang2023evaluatinginducingpersonalitypretrained,cho2025scalingpersonalitycontrolllms,mao2024editingpersonalitylargelanguage}.
Recently, researchers have begun to explore activation intervention techniques \cite{deng2024neuronbasedpersonalitytraitinduction}.
Also, \cite{sun2025personalityvectormodulatingpersonality} propose constructing personality vectors through model fusion.

\section{Methodology}
We first indece a single personality via a system-level prompt, enabling the model to respond to multimodal queries under a fixed personality setting. 
We then extend this approach to multi-personality induction by composing multiple non-conflicting traits. 
Finally, we introduce in-conversation personality switching, allowing the model to dynamically change personality settings across successive turns within a single dialogue.
\autoref{sec:single} formalizes single-personality induction,
\autoref{sec:Multi} introduces multi-personality induction,
and \autoref{sec:switch} describes in-conversation personality switching.
An overview of the pipeline is illustrated in \autoref{fig:methods}.

\begin{figure*}[t]
      \centering
      \includegraphics[width=\linewidth]{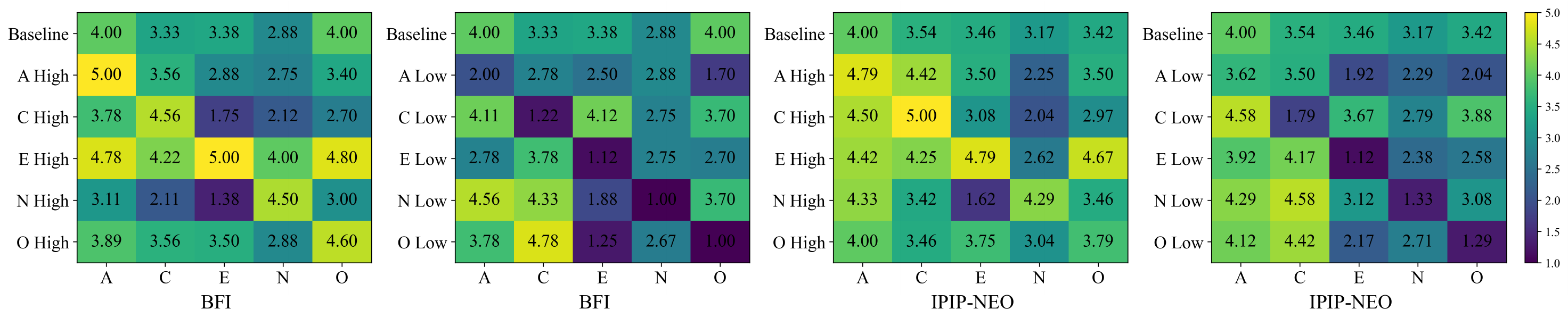}
      \caption{Personality assessment results of the Qwen2.5-VL-7B model based on the BFI and IPIP-NEO.
    We report average trait scores for the Big Five dimensions, including Agreeableness (A), Conscientiousness (C), Extraversion (E), Neuroticism (N), and Openness (O). 
“Baseline” denotes the model without personality conditioning.}
      \label{fig:single}
\end{figure*}

\subsection{Single-Personality Induction}
\label{sec:single}

We model Single-Personality induction as follows.
Let $\pi_{\text{tgt}}$ denote the target model into which personality traits are induced.
$P_{\text{Sys}}^{\text{Per}}$ denotes the prompt used to induce personality and is input into the large language model as the \emph{system prompt}.
The multimodal user query is denoted as
$P_{\text{User}}^{\text{Query}} = \{ Q_{\text{Image}}, Q_{\text{Text}} \}$,
where $Q_{\text{Image}}$ represents the image query and $Q_{\text{Text}}$ represents the textual query.
This multimodal prompt is provided to the model as the \emph{user prompt}.
After constructing both the personality-injection prompt and the query prompt, these components
are passed to the model to obtain the response:
\begin{equation}
\text{Output}
=
\pi_{\text{tgt}}\left(
P_{\text{Sys}}^{\text{Per}},\;
P_{\text{User}}^{\text{Query}}
\right),
\end{equation}
After that, the generated output is evaluated to obtain the final results.

\subsection{Multi-Personality Induction}
\label{sec:Multi}
For multi-personality induction, we merge multiple personality prompts into a single prompt
to induce composite personality behavior. 
Let $\text{Per}_i$ denote the $i$-th Big Five personality trait, including
\emph{Openness}, \emph{Conscientiousness}, \emph{Extraversion}, \emph{Agreeableness}, and \emph{Neuroticism}.
For each Big Five personality dimension, we induce personality only in a single directional tendency and does not simultaneously induce both high- and low-level traits of the same dimension.
For example, within the Openness dimension, either high Openness or low Openness is induced, but not both at the same time, in order to avoid conflicts between opposing personality induction signals.
The resulting multi-personality prompt is defined as:
\begin{equation}
P_{\text{Sys}}^{\text{Per}_{1+2}}
=
\pi_{\text{aux}}\left(
P_{\text{Sys}}^{\text{Per}_1},\;
P_{\text{Sys}}^{\text{Per}_2}
\right)
\end{equation}

In this process, an auxiliary LLM($\pi_{\text{aux}}$) is employed as an agent to combine multiple personality prompts
into a unified representation.
Considering the diversity of possible personality combinations,
we adopt combinations of two personalities (e.g., \emph{Openness}, and \emph{Neuroticism}) and select personality sets with relatively consistent tendencies.
Given that different personalities may exhibit overlapping or even conflicting behavioral
inclinations, we prioritize combinations whose behavioral tendencies are largely aligned during
multi-personality composition. 
This strategy aims to achieve optimal performance while reducing instability caused by personality conflicts.


\begin{table*}[t]
\centering
\small
\setlength{\tabcolsep}{3.5pt}
\begin{tabular}{llccc cc c cc cc cc}
\toprule
\multirow{2}{*}{\textbf{Trait}} &
\multirow{2}{*}{\textbf{Level}} &
\multirow{2}{*}{\textbf{CCBench}} &
\multicolumn{3}{c}{\textbf{Hallusion}} &
\multicolumn{2}{c}{\textbf{MathVista}} &
\multicolumn{2}{c}{\textbf{MMBench}} &
\multicolumn{2}{c}{\textbf{MMMU}} &
\multirow{2}{*}{\textbf{MMStar}} &
\multirow{2}{*}{  \begin{tabular}{c}
    \textbf{SEED-} \\
    \textbf{Bench}
  \end{tabular}}  \\
\cmidrule(lr){4-6}\cmidrule(lr){7-8}\cmidrule(lr){9-10}\cmidrule(lr){11-12}
&
\textbf{} & \textbf{} &
\textbf{aACC} & \textbf{fACC} & \textbf{qACC} &
\textbf{pre} & \textbf{hit} &
\textbf{EN} & \textbf{CN} &
\textbf{val} & \textbf{dev} \\

\midrule

Baseline & -- & 58.2 & \textbf{65.6} & \textbf{36.7} & \textbf{40.6} & 40.2 & 40.6 & \textbf{81.6} & 80.1 & 21.3 & 18.8 & \textbf{61.1} & 77.0 \\

\midrule
\multirow{2}{*}{Agreeableness} & high & 57.5 & 47.9 & 26.9 & 26.6 & 39.8 & 41.2 & 78.9 & 73.5 & 18.0 & 20.0 & 58.7 & 76.1 \\
& low & 56.3 & 63.3 & 30.1 & 36.5 & 42.5 & 40.4 & 77.9 & 76.5 & 27.9 & 23.3 & 60.1 & 76.5 \\

\midrule
\multirow{2}{*}{Conscientiousness} & high & 59.4 & 59.2 & 31.8 & 35.6 & 39.6 & 40.6 & 79.4 & 75.8 & 17.3 & 17.7 & 57.3 & 76.8 \\
& low & 56.5 & \underline{30.1} & \underline{9.24} & \underline{13.0} & \textbf{46.6} & 39.4 & 79.1 & \textbf{81.0} & 31.3 & 29.0 & 61.8 & 77.1 \\

\midrule
\multirow{2}{*}{Extraversion} & high & \underline{47.1} & 35.4 & 11.6 & 14.1 & 39.3 & \underline{38.1} & \underline{75.9} & \underline{57.2} & \underline{16.0} & 14.7 & \underline{51.1} & \underline{72.1} \\
& low & 57.6 & 42.2 & 22.3 & 23.9 & 44.7 & \textbf{43.2} & 79.9 & 80.0 & \textbf{35.8} & \textbf{38.7} & 60.9 & 76.8 \\

\midrule
\multirow{2}{*}{Neuroticism} & high & 55.7 & 34.7 & 15.6 & 16.9 & 39.3 & 40.5 & 78.5 & 75.1 & 18.2 & 17.3 & 58.6 & 76.1 \\
& low & 59.4 & 55.0 & 30.9 & 32.1 & 39.4 & 40.2 & 79.5 & 78.6 & 20.7 & 20.4 & 59.7 & 77.2 \\

\midrule
\multirow{2}{*}{Openness} & high & 56.7 & 52.8 & 27.5 & 31.2 & 37.6 & 40.6 & 78.5 & 74.2 & 17.8 & \underline{16.7} & 57.1 & 76.4 \\
& low & \textbf{62.0} & 59.6 & 32.7 & 34.9 & 41.3 & 40.2 & 79.4 & 80.0 & 25.7 & 22.7 & 60.0 & \textbf{77.2} \\

\bottomrule
\end{tabular}
\caption{Performance on multimodal reasoning benchmarks under different personality conditioning settings. 
Results are reported on CCBench, HallusionBench, MathVista, MMBench, MMMU, MMStar, and SEED-Bench. “Baseline” denotes the model without personality conditioning.}
\label{tab:single-vqa-results}
\end{table*}

\begin{table*}[htbp]
\centering
\small
\setlength{\tabcolsep}{3.5pt}
\begin{tabular}{llcccccccc}
\toprule
\multirow{2}{*}{\textbf{Trait}} &
\multirow{2}{*}{\textbf{Level}} &
\multicolumn{2}{c}{\textbf{BLEU-4}} &
\multicolumn{2}{c}{\textbf{CAPTURE}} &
\multicolumn{2}{c}{\textbf{Objects F1}} &
\multicolumn{2}{c}{\textbf{Relations}} \\
\cmidrule(lr){3-4}\cmidrule(lr){5-6}\cmidrule(lr){7-8}\cmidrule(lr){9-10}
&
\textbf{} &
\textbf{DOCCI} & \textbf{COCO} &
\textbf{DOCCI} & \textbf{COCO} &
\textbf{DOCCI} & \textbf{COCO} &
\textbf{DOCCI} & \textbf{COCO}  \\
\midrule
Baseline & -- & 22.68 & 29.11 & 55.89 & \underline{44.12} & 65.06 & 65.37 & 24.35 & 23.76\\
\midrule
\multirow{2}{*}{Agreeableness} & High & 29.46 &37.89 & 57.52 & 46.14 & 66.54 & 67.81 & 25.51 & 26.24\\
 & Low & 25.50 &33.65 & 56.77 & 45.44 & 65.81 & 66.64 & 25.07 & 25.06  \\
\midrule
\multirow{2}{*}{Conscientiousness} & High & 32.28 & 36.31 & 58.39 & 45.90 & 67.19 &67.77 & 27.77 & 26.69\\
 & Low & \underline{18.05} & \underline{28.35} & \underline{55.24} & 45.02 & \underline{64.37} & 65.76 & \underline{22.45} & 24.25\\
\midrule
\multirow{2}{*}{Extraversion} & High & 32.97 & 37.02 & 57.71 & 45.68 & 66.55 & 67.27 & 26.20 & 25.43\\
 & Low & 26.68 & 31.74 & 56.47 & 44.53 & 65.36 & \underline{65.12} & 25.11 &\underline{23.72}\\
\midrule 
\multirow{2}{*}{Neuroticism} & High & 31.15 &34.67 & 58.37 & 45.40 & 67.65 & 66.76 & 25.72 &24.94\\
 & Low & 30.42 &36.29 & 58.05 &45.42 & 66.88 & 67.12 & 25.92 & 25.39\\
\midrule
\multirow{2}{*}{Openness} & High & 32.13 & 36.41 & 58.36 & 45.97 & 67.28 &67.94 & 26.64 &26.40\\
 & Low & 29.32 &34.81 & 57.61 & 45.82 & 66.88 & 67.27 & 26.08 & 25.14\\
\bottomrule
\end{tabular}
\caption{Image Captioning Performance under Different Personality Conditioning Settings}
\label{tab:qwen-results-caption}
\end{table*}

\subsection{In-conversation Personality Switching}
\label{sec:switch}
In the personality switching setting within a single conversation,
only one personality is induced in a prompt.
Meanwhile, considering the excessively long context resulting from multiple personality switches during the conversation, we will only consider a single personality switch.
In this section, we use $P_i$ to denote the system-level personality
prompt injected at the $i$-th conversation stage.
Specifically, $P_i$ corresponds to $P_{\text{Sys}}^{\text{Per}}$
under the personality setting used at stage $i$.
Starting from an initial conversation state:
$C_0 = \{\}$.

We first induce an initial personality
trait $P_1$ into the model through explicit prompt design. 
After personality initialization, 
we simulate a realistic interaction scenario by engaging in a series
of multi-turn conversations between a simulated user and the model. This process
constructs a conversation context that incorporates both contextual information and
interaction history:
\begin{equation}
C_1 = \{ P_1, (q_1, r_1), (q_2, r_2), \ldots, (q_n, r_n) \},
\end{equation}
where $(q_n, r_n)$ denotes the $n$-th user–model interaction turn.

After completing the initial personality conversation stage, the model is instructed to
switch to a second personality setting $P_2$. The updated conversation context is defined
as: $C_2 = C_1 \oplus P_2$.
where $\oplus$ denotes the operation of appending a new personality induction prompt
to the existing dialogue context.
Subsequently, the model generates responses under the updated context:
\begin{equation}
r_{n+k} = \pi_{\text{tgt}}(C_2, P_{\text{User}}^{\text{quary}}),
\end{equation}
where $P_{\text{User}}^{\text{quary}}$ represents the user query following the personality switch.

Finally, we evaluate the model under the second personality setting using both
personality trait evaluation and multimodal reasoning benchmarks.

\section{Experiment}
In this section, we present our experimental setup in \autoref{sec:setup}. We then report results for single personality (\autoref{sec:single}), multi personality (\autoref{sec:multi}), and personality switching (\autoref{sec:switch}) in multimodal tasks.

\subsection{Setup}
\label{sec:setup}
\textbf{Personality Induction.}\quad
We induce personality via prompting using two methods: \textsc{Personality Prompting} (P\textsuperscript{2})~\cite{jiang2023evaluatinginducingpersonalitypretrained} and \textsc{Big5-Scaler}~\cite{cho2025scalingpersonalitycontrolllms}.
P\textsuperscript{2} converts target traits into structured natural-language descriptions, while \textsc{Big5-Scaler} explicitly specifies Big Five traits with intensity values in the prompt.\\
\textbf{Models.}\quad
We use LLaVA-v1.6-7B~\cite{liu2023visualinstructiontuning} and Qwen2.5-VL-7B~\cite{bai2025qwen25vltechnicalreport} as base models.
\textsc{Big5-Scaler} fails to induce personality in Qwen2.5-VL-7B, and P\textsuperscript{2} fails on LLaVA-v1.6-7B, while P\textsuperscript{2} successfully induces personality in Qwen2.5-VL-7B (see Appendix for detail).
Therefore, our multi-personality and personality-switching analyses focus on Qwen2.5-VL-7B.\\
\textbf{Implementation.}\quad
We evaluate personality expression with two psychometric instruments: the 44-item Big Five Inventory (BFI) and the 120-item IPIP-NEO test, following prior work~\cite{huang2024on,jiang2023evaluatinginducingpersonalitypretrained}.
For multimodal reasoning, we evaluate image captioning on DOCCI500 and COCO-LN500 with BLEU-4, METEOR, and CAPTURE~\cite{zhang2025sccaptionerimprovingimagecaptioning}, and VQA on seven benchmarks: CCBench, HallusionBench, MathVista, MMBench, MMMU, MMStar, and SEED-Bench~\cite{liu2024mmbenchmultimodalmodelallaround,guan2024hallusionbenchadvanceddiagnosticsuite,lu2024mathvistaevaluatingmathematicalreasoning,yue2024mmmumassivemultidisciplinemultimodal,chen2024rightwayevaluatinglarge,li2023seedbenchbenchmarkingmultimodalllms}.

\begin{table*}[t]
\centering
\small
\setlength{\tabcolsep}{3.5pt}
\begin{tabular}{l c ccc cc cc cc c c}
\toprule
\multirow{2}{*}{\textbf{Multi-Trait}} &
\multirow{2}{*}{\textbf{CCBench}} &
\multicolumn{3}{c}{\textbf{Hallusion}} &
\multicolumn{2}{c}{\textbf{MathVista}} &
\multicolumn{2}{c}{\textbf{MMBench}} &
\multicolumn{2}{c}{\textbf{MMMU}} &
\multirow{2}{*}{\textbf{MMStar}} &
\multirow{2}{*}{  \begin{tabular}{c}
    \textbf{SEED-} \\
    \textbf{Bench}
  \end{tabular}}  \\
\cmidrule(lr){3-5}\cmidrule(lr){6-7}\cmidrule(lr){8-9}\cmidrule(lr){10-11}
&  &
\textbf{aACC} & \textbf{fACC} & \textbf{qACC} &
\textbf{pre} & \textbf{hit} &
\textbf{EN} & \textbf{CN} &
\textbf{val} & \textbf{dev} &
\textbf{} & \textbf{} \\
\midrule

$C_{\mathrm{h}} + E_{\mathrm{h}}$ & 58.0 & 45.4 & 23.4 & 24.4 & 36.5 & 38.7 & 78.3 & 64.2 & 20.7 & 17.9 & 48.0 & 73.7 \\
$C_{\mathrm{h}} + O_{\mathrm{l}}$ & 57.8 & 51.2 & 27.2 & 30.5 & 37.3 & 37.4 & 77.9 & 72.8 & 20.7 & 21.3 & 54.1 & 75.9 \\
$E_{\mathrm{h}} + N_{\mathrm{h}}$ & 54.7 & 29.0 & 11.3 & 12.1 & 38.0 & 39.3 & 76.0 & \underline{61.3} & 19.3 & \underline{15.9} & 49.7 & 72.7 \\
$E_{\mathrm{h}} + C_{\mathrm{l}}$ & \underline{44.9} & \underline{27.8} & \underline{9.5}  & \underline{10.8} & 39.5 & \underline{36.8} & \underline{74.3} & 64.6 & 16.8 & 22.0 & 49.9 & 74.5 \\
$C_{\mathrm{l}} + O_{\mathrm{l}}$ & 54.1 & 46.1 & 19.1 & 24.4 & 41.1 & 38.1 & 78.3 & 75.1 & 27.0 & 29.3 & 57.1 & 76.6 \\
$O_{\mathrm{l}} + E_{\mathrm{h}}$ & 57.1 & 41.5 & 19.7 & 21.1 & 38.0 & 39.3 & 78.1 & 66.2 & 21.3 & 20.0 & 52.4 & 73.8 \\
$C_{\mathrm{h}} + E_{\mathrm{l}}$ & 59.0 & 45.0 & 22.8 & 25.1 & 37.4 & 39.2 & 79.0 & 70.0 & 18.4 & 18.7 & 52.4 & 75.6 \\
$N_{\mathrm{h}} + C_{\mathrm{h}}$ & 55.5 & 36.0 & 16.2 & 18.2 & \underline{35.3} & 39.4 & 76.9 & 60.3 & 15.4 & 18.0 & \underline{47.3} & \underline{71.9} \\
$N_{\mathrm{h}} + O_{\mathrm{l}}$ & 59.8 & 35.5 & 15.3 & 18.0 & 36.9 & 38.9 & 77.6 & 63.5 & \underline{15.3} & 18.7 & 50.7 & 73.6 \\

\bottomrule
\end{tabular}
\caption{Performance on multimodal reasoning benchmarks (VQA) under different multi-personality conditioning settings.
In table, C, E, N, and O denote Conscientiousness, Extraversion, Neuroticism, and Openness, respectively. 
The subscripts h and l indicate high and low levels of each trait.
For example, $C_{\mathrm{h}} + E_{\mathrm{h}}$ denotes high Conscientiousness combined with high Extraversion.}
\label{tab:multi-results_vqa}
\end{table*}

\begin{table*}[t]
\centering
\small
\setlength{\tabcolsep}{4pt}
\begin{tabular}{lcccccccc}
\toprule
\multirow{2}{*}{\textbf{Multi-Trait}} &
\multicolumn{2}{c}{\textbf{BLEU-4}} &
\multicolumn{2}{c}{\textbf{CAPTURE}} &
\multicolumn{2}{c}{\textbf{Objects F1}} &
\multicolumn{2}{c}{\textbf{Relations}} \\
\cmidrule(lr){2-3}\cmidrule(lr){4-5}\cmidrule(lr){6-7}\cmidrule(lr){8-9}
&
\textbf{DOCCI} & \textbf{COCO} &
\textbf{DOCCI} & \textbf{COCO} &
\textbf{DOCCI} & \textbf{COCO} &
\textbf{DOCCI} & \textbf{COCO}  \\
\midrule
$C_{\mathrm{h}} + E_{\mathrm{h}}$ & 33.08 & 38.00 & 58.44 & 46.15 & 67.35 & 68.24 & 26.60 & 27.09 \\
$C_{\mathrm{h}} + O_{\mathrm{l}}$ & 32.03 & 37.72 & 58.38 & 45.97 & 67.43 & 67.74 & 26.00 & 25.75 \\
$E_{\mathrm{h}} + N_{\mathrm{h}}$ & 31.21 & 37.93 & 58.28 & 46.10 & 67.72 & 67.90 & 25.11 & 26.28 \\
$E_{\mathrm{h}} + C_{\mathrm{l}}$ & 29.84 & 36.31 & 57.94 & 45.89 & 66.77 & 67.39 & 24.79 & 25.02 \\
$C_{\mathrm{l}} + O_{\mathrm{l}}$ & \underline{26.61} & \underline{36.14} & \underline{57.46} & \underline{45.63} & \underline{66.38} & \underline{67.17} & 25.15 & \underline{24.09} \\
$O_{\mathrm{l}} + E_{\mathrm{h}}$ & 30.70 & 37.68 & 57.97 & 45.83 & 66.94 & 67.60 & \underline{24.75} & 25.95 \\
$C_{\mathrm{h}} + E_{\mathrm{l}}$ & 32.68 & 36.61 & 58.24 & 46.18 & 67.13 & 68.46 & 26.40 & 26.85 \\
$N_{\mathrm{h}} + C_{\mathrm{h}}$ & 34.30 & 38.18 & 58.39 & 46.15 & 67.57 & 68.49 & 26.88 & 28.27 \\
$N_{\mathrm{h}} + O_{\mathrm{l}}$ & 31.65 & 37.34 & 58.18 & 45.66 & 67.45 & 67.73 & 25.23 & 25.63 \\
\bottomrule
\end{tabular}
\caption{Performance of different multi-personality configurations on image captioning.}
\label{tab:multi_personality_caption}
\end{table*}

\begin{figure*}[t]
      \centering
      \includegraphics[width=\linewidth]{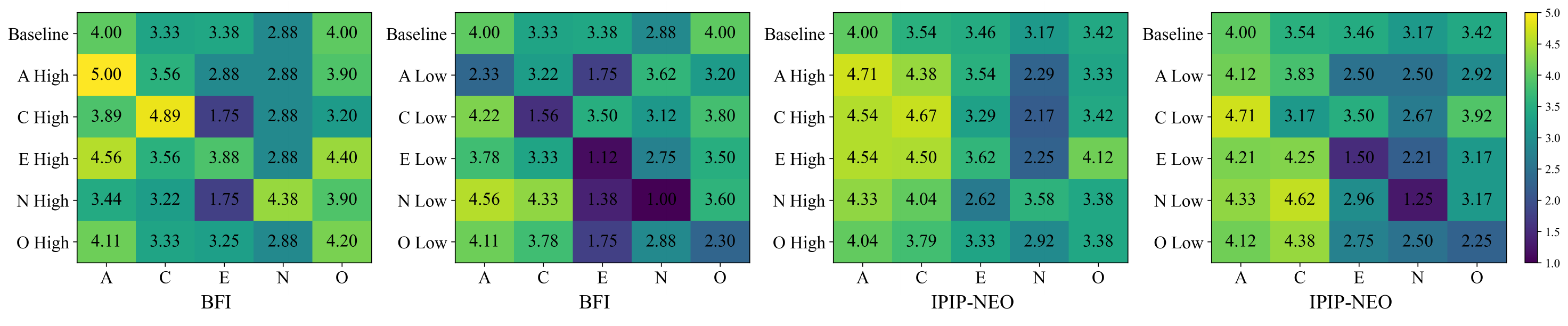}
      \caption{Personality assessment under in-conversation personality switching settings using BFI and IPIP-NEO.}
      \label{fig:switch}
\end{figure*}

\begin{table*}[t]
\centering
\small
\setlength{\tabcolsep}{3.5pt}
\begin{tabular}{llccc cc c cc cc cc}
\toprule
\multirow{2}{*}{\textbf{Trait}} &
\multirow{2}{*}{\textbf{Level}} &
\multirow{2}{*}{\textbf{CCBench}} &
\multicolumn{3}{c}{\textbf{Hallusion}} &
\multicolumn{2}{c}{\textbf{MathVista}} &
\multicolumn{2}{c}{\textbf{MMBench}} &
\multicolumn{2}{c}{\textbf{MMMU}} &
\multirow{2}{*}{\textbf{MMStar}} &
\multirow{2}{*}{  \begin{tabular}{c}
    \textbf{SEED-} \\
    \textbf{Bench}
  \end{tabular}}  \\
\cmidrule(lr){4-6}\cmidrule(lr){7-8}\cmidrule(lr){9-10}\cmidrule(lr){11-12}
&
\textbf{} & \textbf{} &
\textbf{aACC} & \textbf{fACC} & \textbf{qACC} &
\textbf{pre} & \textbf{hit} &
\textbf{CN} & \textbf{EN} &
\textbf{val} & \textbf{dev} \\

\midrule

\multirow{2}{*}{Agreeableness} & high & 55.7 & 49.4 & 28.0 & 28.1 & 39.5 & 39.1 & 77.8 & 71.8 & 20.8 & 20.7 & 55.4 & 75.9 \\
& low & 55.9 & 58.4 & 28.9 & 34.7 & 39.6 & 39.3 & 77.2 & 79.3 & 21.0 & 20.0 & 55.9 & 76.3 \\

\midrule
\multirow{2}{*}{Conscientiousness} & high & 57.6 & 55.7 & 27.7 & 31.9 & 36.6 & 38.4 & 77.4 & 71.5 & 17.1 & 20.0 & 54.1 & 75.8 \\
& low & 57.5 & 53.7 & 26.9 & 30.8 & 39.5 & 39.7 & 77.9 & 75.1 & 21.8 & 23.3 & 56.8 & 76.3 \\

\midrule
\multirow{2}{*}{Extraversion} & high & 54.5 & 52.7 & 28.0 & 29.9 & 38.1 & 39.1 & 77.7 & 63.2 & 19.6 & 20.0 & 50.5 & 73.7 \\
& low & 57.8 & 43.0 & 21.1 & 22.4 & 39.4 & 40.4 & 77.6 & 71.3 & 20.7 & 22.8 & 54.7 & 75.5 \\

\midrule
\multirow{2}{*}{Neuroticism} & high & 53.9 & 48.7 & 25.7 & 26.6 & 38.4 & 39.0 & 77.2 & 66.8 & 19.0 & 18.0 & 50.5 & 73.8 \\
& low & 56.9 & 54.8 & 28.3 & 31.6 & 37.0 & 39.0 & 77.8 & 70.0 & 18.9 & 17.3 & 54.1 & 75.8 \\

\midrule
\multirow{2}{*}{Openness} & high & 55.9 & 57.9 & 28.6 & 33.4 & 38.2 & 39.2 & 77.5 & 64.6 & 17.6 & 18.0 & 50.1 & 73.6 \\
& low & 58.0 & 55.5 & 28.0 & 31.2 & 38.1 & 38.6 & 77.6 & 67.4 & 17.0 & 16.7 & 51.7 & 74.7 \\

\bottomrule
\end{tabular}
\caption{Visual question answering results under In-conversation personality switching.
In the in-conversation personality switching setting, the personality trait shown in the first column denotes the target personality after switching. 
All reported results are evaluated under the post-switch personality condition}
\label{tab:switch-vqa-results}
\end{table*}

\begin{table*}[htbp]
\centering
\small
\setlength{\tabcolsep}{3.5pt}
\begin{tabular}{llcccccccc}
\toprule
\multirow{2}{*}{\textbf{Trait (Switched)}} &
\multirow{2}{*}{\textbf{Level}} &
\multicolumn{2}{c}{\textbf{BLEU-4}} &
\multicolumn{2}{c}{\textbf{CAPTURE}} &
\multicolumn{2}{c}{\textbf{Objects F1}} &
\multicolumn{2}{c}{\textbf{Relations}} \\
\cmidrule(lr){3-4}\cmidrule(lr){5-6}\cmidrule(lr){7-8}\cmidrule(lr){9-10}
&
\textbf{} &
\textbf{DOCCI} & \textbf{COCO} &
\textbf{DOCCI} & \textbf{COCO} &
\textbf{DOCCI} & \textbf{COCO} &
\textbf{DOCCI} & \textbf{COCO}  \\
\midrule

\multirow{2}{*}{Agreeableness} & High & 29.85 &39.25 & 57.41 & 46.50 & 66.40 & 68.61 & 25.43 & 26.77\\
 & Low & 29.39 &38.95 & 57.69 & 46.32 & 66.82 & 68.61 & 25.51 & 25.51  \\
\midrule
\multirow{2}{*}{Conscientiousness} & High & 32.01 & 38.72 & 58.02 & 46.07 & 67.17 &68.46 & 25.88 & 27.46\\
 & Low & 30.63 & 38.39 & 57.94 & 45.87 & 67.12 & 67.97 & 25.39 & 26.24\\
\midrule
\multirow{2}{*}{Extraversion} & High & 31.51 & 38.21 & 57.34 & 45.82 & 66.61 & 67.74 & 25.23 & 26.73\\
 & Low & 31.71 & 37.69 & 57.77 & 45.78 & 67.11 & 67.96 & 25.39 & 26.40 \\
\midrule 
\multirow{2}{*}{Neuroticism} & High & 31.87 &38.47 & 58.15 & 46.26 & 67.53 & 68.90 & 26.32 &26.36\\
 & Low & 30.49 & 38.90 & 57.61 &46.19 & 66.67 & 68.55 & 25.72 & 27.25\\
\midrule
\multirow{2}{*}{Openness} & High & 32.96 & 38.33 & 58.07 & 46.08 & 67.38 &69.87 & 25.68 &26.81\\
 & Low & 31.99 &38.34 & 57.73 & 45.83 & 67.10 & 67.95 & 25.51 & 26.65\\
\bottomrule
\end{tabular}
\caption{Image captioning performance under in-conversation personality switching, evaluated with the switched (post-transition) personality traits.}
\label{tab:qwen-results-switch-caption}
\end{table*}

\subsection{Single-Personality}
\label{sec:single}

\textbf{Personality Trait Assessment.}\quad
\autoref{fig:single} report BFI and IPIP results. 
Prompt-based personality induction effectively shifts the model’s scores along the target dimension, with high settings increasing and low settings decreasing the corresponding trait. 
For instance, under P\textsuperscript{2} on Qwen2.5-VL, Extraversion shows a clear separation (BFI E: 3.38$\rightarrow$5.00 / 1.13).

Besides, we observe systematic cross-trait coupling across settings. Adjusting one dimension often modulates related traits rather than acting independently. 
For example, increasing Extraversion also raises Openness and Agreeableness (E$\uparrow$: BFI O 4.80 and A 4.78 vs baseline O 4.00 and A 4.00), suggesting that induced styles can propagate to correlated personality factors.\\
\textbf{Visual Question Answering.}\quad
\autoref{tab:single-vqa-results} summarizes Qwen2.5-VL performance across multimodal benchmarks under different personality settings. Personality conditioning generally reduces HallusionBench qACC, likely because more expressive or proactive styles conflict with hallucination-sensitive evaluation that rewards conservative, evidence-grounded responses (baseline qACC: 40.6; E$\uparrow$: 14.1). In contrast, lower-intensity settings often lessen this degradation and can even improve performance on standard benchmarks. For example, MMMU improves substantially under low Extraversion (val/dev: 35.8/38.7) compared to high Extraversion (16.0/14.7), consistent with a more restrained style reducing over-commitment under uncertainty.\\
\textbf{Image Captioning.}\quad
\autoref{tab:qwen-results-caption} reports captioning results across BLEU-4, CAPTURE, and object/relation recognition. Generally, personality conditioned models outperform the non-personalized baseline on most metrics, indicating improved descriptive richness and visual–semantic grounding. For example, compared to the baseline BLEU-4 (22.68/29.11 on DOCCI/COCO), the C$\uparrow$ setting yields clear gains (32.28/36.31), alongside consistent improvements on complementary metrics such as CAPTURE and relation recognition. Overall, “positive” trait settings—especially higher Conscientiousness, Extraversion, and Openness—tend to produce more complete and structured captions, benefiting both fluency and visual entity and relationship coverage.
More results about single-personality are in \autoref{single}.

\subsection{Multi-Personality}
\label{sec:multi}
For multi-trait conditioning, we construct personality configurations composed of two traits. The design of these combinations explicitly considers both trait compatibility and trait conflict. Based on the observations from single-trait experiments, we identify several personality traits that have a pronounced impact on task performance. 
For instance, high Conscientiousness and low Openness are found to consistently contribute to improved performance across multiple benchmarks, whereas traits such as Extraversion and Neuroticism exhibit notable modulation effects on reasoning and generation quality. 
Based on these findings, we design a total of nine paired personality configurations.

\textbf{Personality Trait Assessment.}\quad
In most configurations, the personality traits explicitly specified as “high” or “low” still exhibit the corresponding directional trends.
From a global perspective, target trait strengths under multi-trait conditioning are generally comparable to or slightly lower than those achieved with single-trait induction, while remaining consistently above the non-conditioned baseline.

\textbf{Image Captioning and Visual Question Answering.}\quad
\autoref{tab:multi-results_vqa} and \autoref{tab:multi_personality_caption} reports the experimental results of multi-trait conditioning on the visual understanding and question answering tasks and image captioning task.

Multi-trait configurations consistently outperform the non-conditioned baseline across all image captioning metrics, 
demonstrating that incorporating personality information improves both expressive quality and semantic coverage. 
For example, BLEU-4 increases from 22.68/29.11 (DOCCI/COCO) to as high as 34.30/38.18 under multi-trait settings (e.g., $N_{\mathrm{h}} + C_{\mathrm{h}}$), 
while CAPTURE also shows stable gains over the baseline.
Compared with single-trait settings, the advantages of multi-trait configurations primarily arise from the superposition and interaction of multiple personality traits.
For instance, CAPTURE (DOCCI) under multi-trait varies only from 57.46 to 58.44, whereas single-trait spans a wider range (e.g., 55.24–58.39).

As for visual question answering, multi-trait combinations lead to a moderate decline in overall accuracy compared with the non-conditioned baseline or the best single-trait configurations.
For example, HallusionBench qACC decreases from 40.6 to 10.8–30.5, and MMBench-CN drops from 80.1 to 60.3–75.1 under multi-trait settings.
Despite this degradation, the results reveal clear superposition and cancellation effects among personality traits.
For example, certain combinations can offset weaknesses on specific benchmarks (e.g., $C_{\mathrm{l}} + O_{\mathrm{l}}$ improving MMMU val/dev to 27.0/29.3 vs a baseline of 21.3/18.8).
Further details about multi-personality can be found \autoref{app:multi-result}.

\subsection{Personality Switching}
\label{sec:switch}
In the personality switching experiments, the model is first conditioned on an initial personality setting and engages in one round of dialogue under this constraint, after which a second personality is induced and evaluation is conducted under the updated setting.
Considering that switches between conflicting personality are the most representative,  we design the personality switching experiments by sequentially imposing opposite directions of the same personality trait, transitioning from high to low or from low to high within a single trait.

\textbf{Personality Trait Assessment.}\quad
The results in \autoref{fig:switch} indicate that, under multi-turn personality switching, the second-stage personality instruction generally remains effective.
However, compared with single-round personality induction, the strength of the personality expressed after switching is generally reduced.
For instance, switched low Openness reaches BFI O=2.300, noticeably higher than the single-trait $O_{\downarrow}$ case (BFI O=1.000).

\textbf{Image Captioning and Visual Question Answering.}\quad
\autoref{tab:switch-vqa-results} and \autoref{tab:qwen-results-switch-caption} report the performance of personality switching on the image captioning task and visual question answering (VQA) task.
Potentially influenced by both the preceding and subsequent personalities, the personality switching approach outperformed the baseline without personality in image captioning tasks.
Numerically, BLEU-4 increases from 22.68/29.11 (DOCCI/COCO, baseline) to roughly 30.49–32.96 / 37.69–39.25 after switching across traits.
As for VQA: switching mitigates performance degradation observed under certain single-trait settings.
For example, under Extraversion, personality switching raises HallusionBench qACC from 14.1 to above 22.0.
Overall, these results indicate that model behavior after personality switching reflects the joint influence of both previous and current personality constraints. Additional results and analyses for personality switching are provided in \autoref{app:switch}.

\section{Conclusion}
This work systematically studies explicit personality conditioning in MLLMs and introduces an evaluation framework covering single-trait induction, multi-trait composition, and in-conversation personality switching. 
Experiments show that personality induction can improve image captioning, but may hurt performance on precision-heavy tasks such as VQA. 
Under multi-trait and switching settings, we observe balancing and residual effects, suggesting model behavior is jointly influenced by both current and prior personality constraints. 
We also find that prompt-based personality induction transfers poorly from text-only to multimodal settings. Overall, personality control in MLLMs is complex and requires more robust multimodal methods.


\section*{Limitations}

In our experiments, we apply existing text-based personality induction methods to multimodal models, but find that some of them fail to reliably induce the intended traits. Moreover, personality evaluation for multimodal models remains largely text-only, maybe failing to capture multimodal behaviors. These observations highlight the need for dedicated personality injection and evaluation methods tailored to multimodal models.



\bibliography{custom}

\clearpage
\appendix
\section{Experiment Setup}
\label{app:setup}
\subsection{Inducing Personality}
Here, we employ two methods for inducing personality through prompts: 
\textsc{Personality Prompting} (P\textsuperscript{2})~\cite{jiang2023evaluatinginducingpersonalitypretrained} and \textsc{Big5-Scaler}~\cite{cho2025scalingpersonalitycontrolllms}.
P\textsuperscript{2} maps target personality traits (e.g., extraversion, conscientiousness) into structured natural-language descriptions that characterize their typical behavioral tendencies, emotional patterns, and linguistic expression styles. 
\textsc{Big5-Scaler} is a prompt-based personality control method that explicitly encodes Big Five personality traits and their corresponding intensity values into natural-language prompts.

\subsection{Evaluation}
To evaluate personality traits expressed by the model, we adopt two widely used psychometric instruments, namely the Big Five Inventory (BFI) and the IPIP-NEO questionnaire, following prior work. Following \cite{huang2024on}, we use the 44-item BFI test, where each question is presented to the model as a prompt and answered on a 5-point Likert scale ranging from 1 (strongly disagree) to 5 (strongly agree).
For IPIP-NEO test, we employ the 120-item IPIP-NEO test from \cite{jiang2023evaluatinginducingpersonalitypretrained}.

To assess multimodal reasoning capabilities,
we conduct extensive evaluations across Visual Question Answering (VQA) and image captioning tasks.
We follow \cite{zhang2025sccaptionerimprovingimagecaptioning} to evaluate captioning performance,
using DOCCI500 and COCO-LN500 two human-annotated datasets,
and Various metrics including BLEU-4, METEOR, and CAPTURE are used for evaluation.
For VQA aeeseement, we employ 7 benchmarks: (1) Chinese cultural understanding on CCBench \cite{liu2024mmbenchmultimodalmodelallaround}, (2) visual reasoning hallucination detection on HallusionBench \cite{guan2024hallusionbenchadvanceddiagnosticsuite}, (3) visual math reasoning on MathVista \cite{lu2024mathvistaevaluatingmathematicalreasoning}, (4)general reasoning on MMBench \cite{liu2024mmbenchmultimodalmodelallaround}, MMMU \cite{yue2024mmmumassivemultidisciplinemultimodal},
MMStar \cite{chen2024rightwayevaluatinglarge}, and SEED-Bench \cite{li2023seedbenchbenchmarkingmultimodalllms}.

\subsection{Model}
We use LLaVA-v1.6-7B \cite{liu2023visualinstructiontuning} and Qwen2.5-VL-7B \cite{bai2025qwen25vltechnicalreport} as the base models for personality induction.
\autoref{tab:personality_bfi_mpi} and \autoref{tab:personality_bfi_mpi_llava} report the evaluation results on BFI and IPIP.
These results indicate that inducing personality into Qwen2.5-VL-7B using \textsc{Big5-Scaler}~\cite{cho2025scalingpersonalitycontrolllms} was unsuccessful.
Similarly, P\textsuperscript{2}~\cite{jiang2023evaluatinginducingpersonalitypretrained} did not successfully induce personality in LLaVA-v1.6-7B.
However, using P\textsuperscript{2}, we successfully induced personality in the multimodal model Qwen2.5-VL-7B.
Therefore, the following sections on multi-personality and personality switching primarily analyze Qwen2.5-VL-7B.
\section{Single-Personality}
\label{single}
\subsection{Personality Trait Assessment}
\autoref{tab:personality_bfi_mpi} and \autoref{tab:personality_bfi_mpi_llava} show the test results for BFI and IPIP.
Overall, prompt-based induction effectively altered the model’s distribution across target personality dimensions.
The performance trends are consistent across both personality tests. Compared to the absence of personality traits,
high personality settings significantly increase target dimension scores, while low personality settings significantly decrease these scores.
For example, under P\textsuperscript{2} on Qwen2.5-VL, the target trait shows clear separation between high/low settings:
Agreeableness (BFI A: 4.00$\rightarrow$5.00 / 2.00), Conscientiousness (BFI C: 3.33$\rightarrow$4.56 / 1.22),
Extraversion (BFI E: 3.38$\rightarrow$5.00 / 1.13), Neuroticism (BFI N: 2.88$\rightarrow$4.50 / 1.00),
and Openness (BFI O: 4.00$\rightarrow$4.60 / 1.00).

Across all personality settings, significant cross-personality influences are observed beyond changes in the target dimension.
Increasing one personality dimension not only strengthens that dimension itself but also systematically regulates related personality traits.
Agreeableness is typically associated with lower neuroticism and slightly lower extraversion (e.g., A$\uparrow$ yields BFI N 2.75 vs 2.88 baseline, and BFI E 2.88 vs 3.38).
High conscientiousness significantly reduces extraversion and neuroticism (C$\uparrow$: BFI E 1.75, N 2.13), while low conscientiousness is associated with higher extraversion (C$\downarrow$: BFI E 4.13), reflecting reduced self-control.
High extraversion not only enhances expressiveness but also promotes openness and agreeableness (E$\uparrow$: BFI O 4.80 and A 4.78 vs baseline O 4.00 and A 4.00).
High neuroticism is associated with lower conscientiousness and extraversion (N$\uparrow$: BFI C 2.11 and E 1.38), and low openness limits extraversion and expressiveness (O$\downarrow$: BFI E 1.25 and A 3.78).

\subsection{Visual Question Answering}
\autoref{tab:single-vqa-results} summarizes the performance of Qwen2.5-VL on several multimodal evaluation benchmarks under different personality induction settings.

After introducing personality conditioning, HallusionBench scores (qACC) decrease across most settings, likely because personality-induced expressive styles conflict with the conservative requirements of hallucination-sensitive evaluation (baseline qACC: 40.6, while typical high-personality settings drop notably, e.g., A$\uparrow$ 26.6, E$\uparrow$ 14.1, N$_\uparrow$ 16.9).
The reason may be that Hallucination emphasizes conservative reasoning based on visual evidence, while high extraversion, agreeableness, and openness tend to encourage more proactive, elaborate, or imaginative expression, thereby increasing the risk of generating information that is not visually verified, leading to decreased performance.

Conversely, lower-intensity settings often mitigate this degradation compared to their high-intensity counterparts (e.g., A$\downarrow$ qACC 36.5 vs A$\uparrow$ 26.6; N$\downarrow$ 32.1 vs N$\uparrow$ 16.9), consistent with a more restrained response style under uncertainty.
Interestingly, on multiple general multimodal benchmarks (MMBench, MMMU, MMStar, and SEED-Bench), models with lower personality intensity consistently achieve better performance than those with higher personality intensity in several cases; for instance, MMMU improves markedly under low extraversion (val/dev: 35.8/38.7) compared to high extraversion (16.0/14.7), and MMBench-CN is substantially higher under low extraversion (80.0) than high extraversion (57.2).

\subsection{Image Captioning}
\autoref{tab:qwen-results-caption} reports the performance of different personality settings on image captioning tasks, evaluated using BLEU-4, CAPTURE, object recognition, and relation recognition metrics.

Overall, models with personality conditioning generally outperform the non-personalized baseline across most evaluation metrics,
suggesting that personality induction contributes positively to descriptive quality and visual–semantic understanding.
For example, compared to the baseline (BLEU-4: 22.68/29.11 on DOCCI/COCO), several high-trait settings yield large gains, such as C$\uparrow$ (32.28/36.31) and E$\uparrow$ (32.97/37.02), alongside consistent improvements on CAPTURE (baseline 55.89/44.12 vs C$\uparrow$ 58.39/45.90) and relation recognition (baseline 24.35/23.76 vs C$\uparrow$ 27.77/26.69).

Among different personality intensities, settings with more positive traits tend to yield stronger performance.
In particular, higher levels of Conscientiousness, Extraversion, and Openness show consistent improvements on BLEU-4, CAPTURE, and object- and relation-based metrics (e.g., Objects F1 on DOCCI: 65.06 baseline vs 67.19 C$\uparrow$ / 66.55 E$\uparrow$ / 67.28 O$_\uparrow$).
These findings indicate that incorporating appropriate personality can facilitate the generation of more comprehensive descriptions and improve the modeling of visual entities and their relationships.

\begin{table*}[t]
\centering
\small
\setlength{\tabcolsep}{3.5pt}
\begin{tabular}{lllcccccccccc}
\toprule
\multirow{2}{*}{\textbf{Method}} &
\multirow{2}{*}{\textbf{Trait}} &
\multirow{2}{*}{\textbf{Level}} 
& \multicolumn{5}{c}{\textbf{BFI}} 
& \multicolumn{5}{c}{\textbf{IPIP-NEO}} \\
\cmidrule(lr){4-8} \cmidrule(lr){9-13}
& & & A & C & E & N & O & A & C & E & N & O \\
\midrule

& Baseline & -- 
& 4.000 & 3.333 & 3.375 & 2.875 & 4.000
& 4.000 & 3.542 & 3.458 & 3.167 & 3.417 \\

\midrule

\multirow{10}{*}{\textbf{P\textsuperscript{2}}}

& \multirow{2}{*}{Agreeableness}
& High $\uparrow$
& 5.000 & 3.555 & 2.875 & 2.750 & 3.400
& 4.792 & 4.417 & 3.500 & 2.250 & 3.500 \\

&  & Low~~$\downarrow$
& 2.000 & 2.778 & 2.500 & 2.875 & 1.700
& 3.625 & 3.500 & 1.917 & 2.292 & 2.042 \\

\cmidrule(lr){2-13}

& \multirow{2}{*}{Conscientiousness}
& High $\uparrow$
& 3.778 & 4.556 & 1.750 & 2.125 & 2.700
& 4.500 & 5.000 & 3.083 & 2.042 & 2.968 \\

&  & Low~~$\downarrow$
& 4.111 & 1.222 & 4.125 & 2.750 & 3.700
& 4.583 & 1.792 & 3.666 & 2.792 & 3.875 \\

\cmidrule(lr){2-13}

& \multirow{2}{*}{Extraversion}
& High $\uparrow$
& 4.778 & 4.222 & 5.000 & 4.000 & 4.800
& 4.417 & 4.250 & 4.792 & 2.625 & 4.667 \\

&  & Low~~$\downarrow$
& 2.778 & 3.778 & 1.125 & 2.750 & 2.700
& 3.917 & 4.167 & 1.125 & 2.375 & 2.583 \\

\cmidrule(lr){2-13}

& \multirow{2}{*}{Neuroticism}
& High $\uparrow$
& 3.111 & 2.111 & 1.375 & 4.500 & 3.000
& 4.333 & 3.417 & 1.625 & 4.292 & 3.458 \\

&  & Low~~$\downarrow$
& 4.556 & 4.333 & 1.875 & 1.000 & 3.700
& 4.292 & 4.583 & 3.125 & 1.333 & 3.083 \\

\cmidrule(lr){2-13}

& \multirow{2}{*}{Openness}
& High $\uparrow$
& 3.889 & 3.556 & 3.500 & 2.875 & 4.600
& 4.000 & 3.458 & 3.750 & 3.042 & 3.792 \\

&  & Low~~$\downarrow$
& 3.778 & 4.778 & 1.250 & 2.675 & 1.000
& 4.125 & 4.417 & 2.167 & 2.708 & 1.292 \\

\midrule

\multirow{10}{*}{\textbf{\textsc{Big5-Scaler}}}

& \multirow{2}{*}{Agreeableness}
& High $\uparrow$
& 5.000 & 4.000 & 2.625 & 1,250 & 3.100
& 5.000 & 4.875 & 3.125 & 1.375 & 4.083 \\

&  & Low~~$\downarrow$
& 5.000 & 3.667 & 2.125 & 1.375 & 3.000
& 4.958 & 4.458 & 2.791 & 1.916 & 3.375 \\

\cmidrule(lr){2-13}

& \multirow{2}{*}{Conscientiousness}
& High $\uparrow$
& 4.000 & 5.000 & 2.500 & 1.500 & 2.100 & 4.791 & 5.000 & 3.042 & 1.292 & 3.292 \\

&  & Low~~$\downarrow$
& 3.444 & 5.000 & 2.125 & 2.500 & 1.800 & 4.416 & 5.000 & 2.542 & 1.583 & 2.833 \\

\cmidrule(lr){2-13}

& \multirow{2}{*}{Extraversion}
& High $\uparrow$
& 4.111 & 3.111 & 5.000 & 2.500 & 3.300 & 4.250 & 4.292 & 4.792 & 2.000 & 4.292 \\

&  & Low~~$\downarrow$
& 3.444 & 2.889 & 5.000 & 2.500 & 3.100 & 4.375 & 4.083 & 4.792 & 2.042 & 4.042 \\

\cmidrule(lr){2-13}

& \multirow{2}{*}{Neuroticism}
& High $\uparrow$
& 2.777 & 1.889 & 2.125 & 5.000 & 2.100 & 3.833 & 2.208 & 1.833 & 4.500 & 3.333 \\

&  & Low~~$\downarrow$
& 2.778 & 1.555 & 2.125 & 5.000 & 2.400 & 3.833 & 2.583 & 2.000 & 4.250 & 3.208 \\

\cmidrule(lr){2-13}

& \multirow{2}{*}{Openness}
& High $\uparrow$
& 3.888 & 3.333 & 4.500 & 3.250 & 5.000 & 4.500 & 4.625 & 4.000 & 1.833 & 4.833 \\

&  & Low~~$\downarrow$
& 3.777 & 2.889 & 2.625 & 3.625 & 5.000 & 4.583 & 3.917 & 3.625 & 2.250 & 4.625 \\

\bottomrule
\end{tabular}
\caption{Personality assessment results of the Qwen2.5-VL-7B model based on the BFI and IPIP-NEO under different personality induction methods.
We report average trait scores for the Big Five dimensions, including Agreeableness (A), Conscientiousness (C), Extraversion (E), Neuroticism (N), and Openness (O). 
“Baseline” denotes the model without personality conditioning.}
\label{tab:personality_bfi_mpi}
\end{table*}

\begin{table*}[t]
\centering
\small
\setlength{\tabcolsep}{3.5pt}
\begin{tabular}{llcccccccccc}
\toprule
\multirow{2}{*}{\textbf{Trait}} & \multirow{2}{*}{\textbf{Level}} 
& \multicolumn{5}{c}{\textbf{BFI}} 
& \multicolumn{5}{c}{\textbf{IPIP-NEO}} \\
\cmidrule(lr){3-7} \cmidrule(lr){8-12}
&  & A & C & E & N & O & A & C & E & N & O \\
\midrule

Baseline
& -- 
& 4.556 & 4.556 & 3.875 & 2.000 & 3.900
& 3.167 & 4.208 & 4.250 & 2.708 & 3.958 \\

\midrule

\multirow{2}{*}{Agreeableness}
& High $\uparrow$
& 4.222 & 3.889 & 3.625 & 3.500 & 4.200
& 2.292 & 3.583 & 4.167 & 3.375 & 3.333 \\
& Low~~$\downarrow$
& 2.222 & 3.556 & 2.375 & 2.000 & 3.000
& 2.542 & 3.708 & 3.375 & 2.708 & 3.875 \\

\midrule

\multirow{2}{*}{Conscientiousness}
& High $\uparrow$
& 3.666 & 4.222 & 3.750 & 3.125 & 4.000
& 2.292 & 3.708 & 3.875 & 3.333 & 3.583 \\
& Low~~$\downarrow$
& 3.222 & 3.556 & 3.125 & 3.500 & 4.200
& 2.167 & 2.958 & 4.000 & 3.708 & 3.125 \\

\midrule

\multirow{2}{*}{Extraversion}
& High $\uparrow$
& 4.222 & 4.556 & 4.625 & 2.750 & 4.500
& 2.667 & 3.708 & 4.125 & 3.083 & 3.875 \\
& Low~~$\downarrow$
& 3.222 & 2.889 & 2.750 & 3.875 & 4.000
& 3.000 & 3.625 & 3.417 & 3.458 & 3.542 \\

\midrule

\multirow{2}{*}{Neuroticism}
& High $\uparrow$
& 2.889 & 3.222 & 3.500 & 4.625 & 3.900
& 2.417 & 2.833 & 3.625 & 4.458 & 4.125 \\
& Low~~$\downarrow$
& 3.333 & 3.556 & 3.750 & 3.500 & 4.400
& 2.208 & 3.458 & 4.083 & 3.333 & 3.333 \\

\midrule

\multirow{2}{*}{Openness}
& High $\uparrow$
& 3.889 & 3.889 & 4.250 & 3.500 & 4.200
& 2.292 & 3.458 & 4.000 & 3.458 & 3.750 \\
& Low~~$\downarrow$
& 3.000 & 4.222 & 3.125 & 3.500 & 4.200
& 2.208 & 3.458 & 4.083 & 3.458 & 3.125 \\

\bottomrule
\end{tabular}
\caption{Personality assessment results of the LLaVa-v1.6-7B model based on the BFI and IPIP-NEO under different personality induction methods.}
\label{tab:personality_bfi_mpi_llava}
\end{table*}

\section{Multi-Personality}
\label{app:multi-result}

\begin{table*}[t]
\centering
\small
\setlength{\tabcolsep}{3.5pt}
\begin{tabular}{lcccccccccc}
\toprule
\textbf{Multi-Trait} 
& \multicolumn{5}{c}{\textbf{BFI}} 
& \multicolumn{5}{c}{\textbf{IPIP-NEO}} \\
\cmidrule(lr){2-6} \cmidrule(lr){7-11}
 & A & C & E & N & O & A & C & E & N & O \\
\midrule

$C_{\mathrm{h}} + E_{\mathrm{h}}$ & 3.889 & 4.889 & 4.500 & 2.500 & 3.800 & 4.583 & 5.000 & 4.250 & 1.583 & 3.750 \\
$C_{\mathrm{h}} + O_{\mathrm{l}}$ & 3.111 & 5.000 & 1.625 & 1.875 & 1.000 & 4.500 & 4.750 & 2.292 & 1.708 & 1.167 \\
$E_{\mathrm{h}} + N_{\mathrm{h}}$ & 3.111 & 1.667 & 4.500 & 4.500 & 3.800 & 4.667 & 3.625 & 4.500 & 3.792 & 4.583 \\
$E_{\mathrm{h}} + C_{\mathrm{l}}$ & 4.667 & 1.444 & 4.500 & 2.875 & 4.500 & 4.292 & 1.875 & 4.083 & 2.625 & 4.667 \\
$C_{\mathrm{l}} + O_{\mathrm{l}}$ & 3.444 & 1.667 & 2.500 & 1.875 & 1.100 & 4.250 & 3.125 & 2.250 & 2.708 & 1.417 \\
$O_{\mathrm{l}} + E_{\mathrm{h}}$ & 4.222 & 4.778 & 4.500 & 2.500 & 1.100 & 4.833 & 4.458 & 3.667 & 2.417 & 1.708 \\
$C_{\mathrm{h}} + E_{\mathrm{l}}$ & 2.444 & 4.556 & 1.125 & 1.875 & 2.000 & 3.833 & 4.792 & 1.708 & 1.625 & 2.083 \\
$N_{\mathrm{h}} + C_{\mathrm{h}}$ & 3.111 & 4.667 & 1.375 & 4.500 & 2.100 & 4.375 & 4.625 & 2.167 & 3.417 & 2.625 \\
$N_{\mathrm{h}} + O_{\mathrm{l}}$ & 2.444 & 4.667 & 1.250 & 4.375 & 1.000    & 4.042 & 4.208 & 1.500 & 3.375 & 1.250 \\

\bottomrule
\end{tabular}
\caption{BFI and IPIP-NEO scores under different multi-personality configurations.}
\label{tab:multi_personality_qwen}
\end{table*} 
\subsection{Personality Trait Assessment}
\label{app:multi-person}

We can know that the model is capable of jointly modeling multiple personality traits even with simple prompt-based conditioning.
For example, in $C_{\mathrm{h}} + E_{\mathrm{h}}$, both traits move upward (BFI C=4.889, E=4.500 vs baseline C=4.556, E=3.875), while in $C_{\mathrm{h}} + O_{\mathrm{l}}$ the openness is strongly suppressed (BFI O=1.000 vs baseline 3.900) with elevated conscientiousness (BFI C=5.000). Similarly, $E_{\mathrm{h}} + N_{\mathrm{h}}$ simultaneously increases extraversion and neuroticism (BFI E=4.500, N=4.500 vs baseline E=3.875, N=2.000), confirming joint controllability across traits.

Meanwhile, the strength of the target traits under multi-trait settings is generally lower than or comparable to that observed under single-trait induction, while remaining consistently higher than that of the non-conditioned baseline. Concretely, some “high” targets become slightly moderated compared with single-trait peaks (e.g., multi-trait $E_{\mathrm{h}}$ often reaches BFI E=4.500, compared to single-trait $E_{\uparrow}$ at 5.000), while still staying above baseline E=3.875; likewise, “low” traits remain strongly suppressed (e.g., $O_{\mathrm{l}}$ stays near BFI O=1.000–1.100 vs baseline 3.900). This suggests that multi-trait conditioning does not undermine personality controllability, but instead introduces a balancing effect, where the expression of individual traits is moderated by the presence of additional personality constraints.

\subsection{Image Captioning}
\label{app:multi-caption}
Multi-trait configurations significantly outperform the non-conditioned baseline across all evaluation metrics, indicating that incorporating personality information effectively enhances both the expressive quality and semantic coverage of generated captions. Numerically, BLEU-4 improves from 22.68/29.11 (DOCCI/COCO) to 26.61–34.30 / 36.14–38.18 across multi-trait settings, while CAPTURE rises from 55.89/44.12 to a tight band of 57.46–58.44 / 45.63–46.18, showing both gains and high stability. In most cases, multi-trait settings further amplify this performance gain (e.g., the best BLEU-4 appears in $N_{\mathrm{h}} + C_{\mathrm{h}}$: 34.30/38.18).

Compared with single-trait configurations, multi-trait combinations achieve comparable or slightly better results on most metrics, with notably higher stability on CAPTURE and relation-related measures. For instance, CAPTURE(DOCCI) under multi-trait varies only from 57.46 to 58.44, whereas single-trait spans a wider range (e.g., 55.24–58.39). On relation recognition, several multi-trait settings are consistently strong (COCO Relations up to 28.27 in $N_{\mathrm{h}} + C_{\mathrm{h}}$, compared to 23.76 baseline). A closer examination of different personality combinations shows that configurations involving high Conscientiousness ($C_{\mathrm{h}}$) are generally more robust (e.g., $C_{\mathrm{h}} + E_{\mathrm{h}}$ and $C_{\mathrm{h}} + E_{\mathrm{l}}$ both achieve high BLEU-4 $\ge$32.68 on DOCCI). In addition, although low Openness may suppress performance when applied in isolation, it can still maintain relatively high performance in certain multi-trait combinations (e.g., $C_{\mathrm{h}} + O_{\mathrm{l}}$ achieves 32.03/37.72 BLEU-4 and 58.38/45.97 CAPTURE), suggesting the presence of compensatory interactions among personality traits.

\subsection{Visiual Question Answering}
\label{app:multi-vqa}
From an overall perspective, multi-trait combinations exhibit a moderate decrease in overall accuracy compared with the non-conditioned baseline or the best-performing single-trait configurations. For example, HallusionBench qACC drops from 40.6 (baseline) to 10.8–30.5 across multi-trait settings, and MMBench-CN decreases from 80.1 to 60.3–75.1 in most combinations. At the same time, the results indicate the presence of superposition effects among personality traits. In particular, the high Extraversion ($E_{\mathrm{h}}$) configuration, which performs the worst under single-trait conditioning, shows a noticeable mitigation of performance degradation when combined with other personality traits: compared to single-trait $E_{\uparrow}$ (CCBench 47.1; Hallusion qACC 14.1), pairing it with other traits raises CCBench to 57.1–58.0 in $O_{\mathrm{l}} + E_{\mathrm{h}}$ / $C_{\mathrm{h}} + E_{\mathrm{h}}$ and improves Hallusion qACC to 21.1–24.4, indicating partial recovery.

Furthermore, the performance variations observed under multi-trait settings also reflect a degree of cancellation effect among different personality dimensions. For instance, while several combinations reduce general benchmark scores (e.g., MMStar from 61.1 baseline down to 47.3–54.1 in some settings), other configurations can counteract weaknesses on specific tasks: $C_{\mathrm{l}} + O_{\mathrm{l}}$ yields MMMU val/dev 27.0/29.3 (vs baseline 21.3/18.8), suggesting that trait interactions can redistribute performance across benchmarks rather than uniformly shifting it downward.

\section{Personality Switching}
\label{app:switch}
\begin{table*}[t]
\centering
\small
\setlength{\tabcolsep}{3.5pt}
\begin{tabular}{llcccccccccc}
\toprule
\multirow{2}{*}{\textbf{Trait}} & \multirow{2}{*}{\textbf{Level}} 
& \multicolumn{5}{c}{\textbf{BFI}} 
& \multicolumn{5}{c}{\textbf{IPIP-NEO}} \\
\cmidrule(lr){3-7} \cmidrule(lr){8-12}
&  & A & C & E & N & O & A & C & E & N & O \\
\midrule

\multirow{2}{*}{Agreeableness}
& High $\uparrow$
& 5.000  & 3.556 & 2.875 & 2.875 & 3.900 & 4.708 & 4.375 & 3.542 & 2.292 & 3.333 \\
& Low~~$\downarrow$
& 2.333  & 3.222 & 1.750 & 3.625 & 3.200 & 4.125 & 3.833 & 2.500 & 2.500 & 2.917 \\

\midrule

\multirow{2}{*}{Conscientiousness}
& High $\uparrow$
& 3.889  & 4.889 & 1.750 & 2.875 & 3.200 & 4.542 & 4.667 & 3.292 & 2.167 & 3.417 \\
& Low~~$\downarrow$
& 4.222  & 1.556 & 3.500 & 3.125 & 3.800 & 4.708 & 3.167 & 3.500 & 2.667 & 3.917 \\

\midrule

\multirow{2}{*}{Extraversion}
& High $\uparrow$
& 4.556  & 3.556 & 3.875 & 2.875 & 4.400 & 4.542 & 4.500 & 3.625 & 2.250 & 4.125 \\
& Low~~$\downarrow$
& 3.778  & 3.333 & 1.125 & 2.750 & 3.500 & 4.208 & 4.250 & 1.500 & 2.208 & 3.167 \\

\midrule

\multirow{2}{*}{Neuroticism}
& High $\uparrow$
& 3.444  & 3.222 & 1.750 & 4.375 & 3.900 & 4.333 & 4.042 & 2.625 & 3.583 & 3.375 \\
& Low~~$\downarrow$
& 4.556  & 4.333 & 1.375 & 1.000 & 3.600 & 4.333 & 4.625 & 2.958 & 1.250 & 3.167 \\

\midrule

\multirow{2}{*}{Openness}
& High $\uparrow$
& 4.111  & 3.333 & 3.250 & 2.875 & 4.200 & 4.042 & 3.792 & 3.333 & 2.917 & 3.375 \\
& Low~~$\downarrow$
& 4.111  & 3.778 & 1.750 & 2.875 & 2.300 & 4.125 & 4.375 & 2.750 & 2.500 & 2.250 \\

\bottomrule
\end{tabular}
\caption{Personality assessment scores under in-conversation personality switching settings, evaluated using BFI and IPIP-NEO.}
\label{tab:switch-personality_bfi_mpi}
\end{table*}

\subsection{Personality Trait Assessment}
\label{app:switch-personality}
\autoref{tab:switch-personality_bfi_mpi} reports the personality assessment results after personality switching.
The results indicate that, under most simple multi-turn dialogue settings, the personality injected in the second stage remains effective.
Concretely, the switched target traits still show clear directional separation, e.g., Neuroticism (BFI N: 4.375 vs 1.000), Extraversion (BFI E: 3.875 vs 1.125), and Conscientiousness (BFI C: 4.889 vs 1.556), indicating the second-stage instruction is largely followed.
This suggests that the model is capable of dynamically updating its behavioral characteristics in response to the most recent personality instruction, rather than being fixed to the previously induced personality state.
However, compared with single-round personality induction, the strength of the personality expressed after switching is generally reduced.
For example, Openness under switching shows: O$\downarrow$ yields BFI O=2.300, which is substantially higher (less suppressed) than the single-trait O$\downarrow$ case (BFI O=1.000), and Agreeableness low is also less extreme (BFI A=2.333 vs 2.000 under single-trait A$_\downarrow$).
This attenuation is likely due to residual effects from the opposite personality applied in the previous dialogue turn, which partially interferes with the explicit expression of the newly injected personality.
These findings imply that personality switching in multi-turn interactions is not entirely memory-free, and prior personality states may exert a lasting influence on subsequent personality modeling.

\subsection{Image Captioning}
\label{switch_caption}
\autoref{tab:qwen-results-switch-caption} reports the performance of personality switching on the image captioning task.
It can be observed that different personality switching settings consistently lead to performance improvements compared to the non-conditioned baseline.
Numerically, BLEU-4 increases from 22.68/29.11 (DOCCI/COCO, baseline) to roughly 30.49–32.96 / 37.69–39.25 after switching across traits, with CAPTURE improving from 55.89/44.12 to 57.34–58.15 / 45.78–46.50.
Moreover, for the same personality dimension, the two opposite switching directions yield relatively similar performance.
For instance, Agreeableness switching yields close BLEU-4 (29.85 vs 29.39 on DOCCI; 39.25 vs 38.95 on COCO) and CAPTURE (57.41 vs 57.69 on DOCCI; 46.50 vs 46.32 on COCO), and similarly for Extraversion (BLEU-4 DOCCI 31.51 vs 31.71; COCO 38.21 vs 37.69).
This suggests that during personality switching, the model’s generation behavior may be jointly influenced by both the preceding and the subsequent personality settings, resulting in comparable outcomes.

\subsection{Visual Question Answering}
\label{app:switch-vqa}
\autoref{tab:switch-vqa-results} reports the performance of personality switching on the visual question answering (VQA) task.
Similarly, a balancing effect between the preceding and subsequent personality settings can also be observed in the VQA results,
where many switched settings fall between the two single-trait extremes. For example, under Extraversion, single-trait $E_{\uparrow}$ performs poorly on CCBench (47.1), while $E_{\downarrow}$ is higher (57.6); after switching, performance is partially recovered (54.5–57.8). On HallusionBench qACC, switching also mitigates the severe degradation of single-trait $E_{\uparrow}$ (14.1), reaching 22.4–29.9 after switching.
indicating that the model’s behavior after switching reflects the combined influence of both personality conditions rather than being dominated by a single trait.

This observation suggests that the model behavior after personality switching is not solely governed by the newly injected personality, but is instead jointly influenced by both the previous and the current personality constraints, resulting in an intermediate performance regime between positive and negative personality settings.

\end{document}